\documentclass[11pt,a4paper]{article}
\usepackage[hyperref]{acl2021}
\usepackage{times}
\usepackage{latexsym}
\usepackage{color}
\usepackage{multirow}
\usepackage{forest}
\usepackage{tikz}
\usepackage{algorithm}
\usepackage{algpseudocode}
\usepackage{amsmath}
\usepackage{graphics}
\usepackage{subfigure}
\usepackage{mwe}
\usepackage{amssymb}
\usepackage{bm}
\usepackage{amsmath}
\usepackage{makecell}
\usepackage{tabularx}
\usepackage{url}
\usepackage{pgfplots}
\usepackage{CJKutf8}
\pgfplotsset{width=10cm,compat=1.9}
\usepackage{booktabs}
\usepackage{dsfont}
\usepackage{arydshln}
\usepackage{soul}

\usepackage{microtype}

\aclfinalcopy 
\def\aclpaperid{3357}

\definecolor{electricviolet}{rgb}{0.56, 0.0, 1.0}

\newcommand{\xuebo}[1]{{\color{black} #1}}

\newcommand{\oracle}{\textsc{Oracle}\xspace}
\newcommand{\random}{\textsc{Random}\xspace}
\newcommand{\pre}{\textsc{Pretrained}\xspace}

\usetikzlibrary{arrows.meta,calc,decorations.markings,math,arrows.meta,patterns}
\usetikzlibrary{decorations.pathreplacing}
\usetikzlibrary{matrix}

\title{On the Copying Behaviors of Pre-Training \\ for Neural Machine Translation}
  
\author{Xuebo Liu$^{1}$\thanks{~~Work was done when Xuebo Liu and Liang Ding were interning at Tencent AI Lab.} , Longyue Wang$^{2}$, Derek F. Wong$^{1}$, Liang Ding$^{3}$, \\ \bf Lidia S. Chao$^{1}$, Shuming Shi$^{2}$ and Zhaopeng Tu$^{2}$\\
  $^{1}$NLP$^2$CT Lab, Department of Computer and Information Science, 
  University of Macau \\
  $^{2}$Tencent AI Lab $^{3}$The University of Sydney\\
  {\tt nlp2ct.xuebo@gmail.com, \{derekfw,lidiasc\}@um.edu.com} \\
  {\tt \{vinnylywang,shumingshi,zptu\}@tencent.com} \\
  {\tt ldin3097@sydney.edu.au}}

\date{}

\begin{document}
\maketitle
\begin{abstract}
Previous studies have shown that initializing neural machine translation (NMT) models with the pre-trained language models (LM) can speed up the model training and boost the model performance.
In this work, we identify a critical side-effect of pre-training for NMT, which is due to the discrepancy between the training objectives of LM-based pre-training and NMT.
\xuebo{Since the LM objective learns to reconstruct a few source tokens and copy most of them, the pre-training initialization would affect the copying behaviors of NMT models.
We provide a quantitative analysis of copying behaviors by introducing a metric called {\it copying ratio}, which empirically shows that pre-training based NMT models have a larger copying ratio than the standard one.}
In response to this problem, we propose a simple and effective method named {\it copying penalty} to control the copying behaviors in decoding.
Extensive experiments on both in-domain and out-of-domain benchmarks show that the copying penalty method consistently improves translation performance by controlling copying behaviors for pre-training based NMT models.
Source code is freely available at \url{https://github.com/SunbowLiu/CopyingPenalty}.
\end{abstract}

\section{Introduction}
Self-supervised pre-training~\cite{Devlin:2018uk,song2019mass}, which acquires general knowledge from a large amount of unlabeled data to help {\em better} and {\em faster} learning downstream tasks, has an intuitive appeal for neural machine translation (\citealp[NMT;][]{bahdanau2014neural,DBLP:journals/corr/VaswaniSPUJGKP17}).
One direct way to utilize pre-trained knowledge is initializing the NMT model with a pre-trained language model (LM) before training it on parallel data~\cite{lample2019cross,Liu:2020mbart}.
As a range of surface, syntactic and semantic information has been encoded in the initialized parameters~\citep{jawahar-etal-2019-bert,goldberg2019assessing}, they are expected to bring benefits to NMT models and hence the translation quality.

\begin{table}[t]
    \centering
    \scalebox{0.98}{
    \begin{tabular}{p{1cm}p{5.9cm}}
    \toprule
    \multicolumn{2}{l}{\em LM Pre-Training: $\mathcal{L}_\mathrm{PT}=-\log P(\bf{x}|\widetilde{\bf{x}})$} \\ \hdashline
        \bf Source & Military \underline{{ }{ }{ }} Field Marshal Hussein \underline{{ }{ }{ }} \underline{{ }{ }{ }} in attendance. \\
        \bf Target & \hl{Military} ruler \hl{Field} \hl{Marshal} \hl{Hussein} Tantawi was \hl{in} \hl{attendance}. \\
                \midrule
\multicolumn{2}{l}{\em NMT Training: $\mathcal{L}_\mathrm{NMT}=-\log P(\bf{y}|\bf{x})$} \\      \hdashline     
        \bf Source & Military ruler Field Marshal Hussein Tantawi was in attendance. \\
        \bf Target & Der Militärführer Feldmarschall \hl{Hussein} \hl{Tantawi} war anwesend. \\
    \bottomrule
    \end{tabular}}
    \caption{Training objective gap between Seq2Seq LM pre-training and NMT training. LM learns to reconstruct a few source tokens and copy most of them, while NMT learns more translation rather than copying. \underline{Underlines} denote artificial noises, and \hl{highlights} indicate expected copying tokens.}
    \label{tab:intro_example}
\end{table}

However, there is a discrepancy between the training objective of sequence-to-sequence LM pre-training and that of NMT training. 
As shown in Table~\ref{tab:intro_example}, LM learns to reconstruct all source tokens with some noises, while NMT learns to translate most source tokens and copy few of them. 
\citet{knowles2018context} and \citet{Liu:2020mbart} show that LM pre-training requires to copy $\sim$65\% of tokens, while NMT training only needs to copy $<$10\%. 
We believe that unexpected knowledge can be propagated to the NMT model via pre-training, which may bias NMT models to mistakenly copy source tokens to the target side.
For example, the source word ``Field Marshal'' might be mistakenly copied to the target side by pre-training based NMT models, since such copying behaviors can be learned in the pre-training stage.

\xuebo{In this paper, we first validate the change of copying behaviors in NMT models initialized with the pre-training weights. To this end, we propose a metric named {\it copying ratio} to quantitatively measure the extent of copying behaviors of NMT models.}
Experimental results on the WMT14 En-De data show that the NMT model with pre-training improves translation performance at the cost of introducing more copying predictions.
Analyses on model training show that the NMT model with pre-training attempts to forget the copying behaviors transferred from pre-training, while the vanilla NMT model learns in the opposite way. 
\xuebo{Due to the dominating copying behaviors in the pre-training, the copying ratio of pre-training based NMT model (i.e., 10.8\%) is much higher than that of the vanilla NMT model (i.e., 9.3\%).}
Extensive analyses show that higher copying ratios severely hurt sentence fluency and word accuracy in translations, particularly for the translation of proper nouns, \xuebo{establishing the necessity for controlling the copying behaviors of NMT models.}

To tackle this problem, we propose a simple and effective {\it copying penalty} to control the copying behaviors in inference, which requires no modification to model architectures and training algorithms.
Specifically, we introduce a new regularizing term to the prediction at each time step, which guides the model to copy source tokens only when the model is highly confident.
Experimental results on the WMT14 English-German and the OPUS German-English benchmark demonstrate that the proposed approach can significantly control copying behaviors in NMT models, making the model more accurately generate copying tokens.

Our contributions are summarized as follows:
\begin{itemize}
    \item We reveal a critical side-effect of pre-training for NMT, where pre-training introduces more copying behaviors into NMT outputs.
    \item We propose a simple and effective {\em copying penalty} to further improve the performance of NMT models with pre-training by controlling copying behaviors in generated translation.
    \item \xuebo{We find that the domain containing a large number of copying tokens (e.g., the IT) benefits more from the proposed copying penalty.}
\end{itemize}

\section{Observing Copying Behavior Changes}
The fact is that some source words are excessively copied by NMT models from the source to the target side instead of being translated, which leads to a high copying ratio in NMT outputs.
In this section, we first propose a metric to measure the copying ratio of model predictions. 
Second, we quantitatively investigate the effect of pre-training on NMT in the perspective of copying behaviors.
We expect to provide more evidence for controlling the copying behaviors of NMT models.

\subsection{Experimental Setup}

\paragraph{Data}
We conducted experiments on the widely-used WMT14 English-German benchmark. 
We used the processed data provided by~\citet{DBLP:journals/corr/VaswaniSPUJGKP17}, which consists of 4.5M sentence pairs.\footnote{\url{https://drive.google.com/uc?id=0B_bZck-ksdkpM25jRUN2X2UxMm8}} 
We used all the training data for model training. 
The validation set is newstest2013 of 3,000 examples and the test set is newstest2014 of 3,003 examples.

\paragraph{Models and Settings}
We implemented all the models by the open-sourced toolkit {\tt fairseq}~\citep{ott2019fairseq}.\footnote{\url{https://github.com/pytorch/fairseq}}
We used 8 V100 GPUs for the experiments.
We mainly compared two models: 1) \random, which is a vanilla NMT model whose weights are randomly initialized without pre-training; and 2) \pre, an NMT model using the weights of pre-trained {\tt mBART.cc25}\footnote{\url{https://github.com/pytorch/fairseq/tree/master/examples/mbart}} for parameter initialization, which has shown its usability and reliability for translation tasks~\citep{tran2020cross,tang2020multilingual}.

For the training of \random, we used the Transformer {\it big} setting of~\citet{ott2018scaling} with a huge training batch size of 460K tokens.\footnote{\url{https://github.com/pytorch/fairseq/blob/master/examples/scaling_nmt/README.md\#3-train-a-model}}
For \pre, we fine-tuned on the pre-trained {\tt mBART.cc25} with a training batch size of 131K tokens. 
The hyperparameters keep the same with \random except the 0.2 label smoothing, 2500 warm-up steps, and 1e-4 maximum learning rate. 

\paragraph{Evaluation}
For each model, we selected the checkpoint with the lowest perplexity on the validation set for testing.
The beam size is 5 and the length penalty is 0.6.
In addition to reporting the commonly-used 4-gram BLEU score~\citep{papineni2002bleu}, we also report Translation Error Rate (TER)~\citep{snover2006study} to better capture the translation performance of unigrams, which more directly reflects the copying behaviors of NMT models.
Both the scores are calculated by \texttt{sacrebleu}~\citep{post-2018-call} with de-tokenized text and unmodified references.\footnote{BLEU+c.mixed+\#.1+s.exp+tok.13a+v.1.4.14}$^,$\footnote{TER+lang.en-de+tok.tercom-nonorm-punct-noasian-uncased+version.1.4.14}

\begin{table}[t]
    \centering
    \scalebox{0.84}{
    \begin{tabular}{lp{5.9cm}}
    \toprule
        \bf Source & Military ruler Field \textcolor{red}{\it Marshal} \textcolor{blue}{Hussein Tantawi} was in attendance. \\
        \bf Target & Der Militärführer Feldmarschall \textcolor{blue}{Hussein Tantawi} war anwesend. \\
                \midrule
        \bf \random & Anwesend war der Militärmachthaber Feldmarschall \textcolor{blue}{Hussein Tantawi}. \\
        \bf \pre & Militärischer Feldherr \textcolor{red}{\it Marshal} \textcolor{blue}{Hussein Tantawi} war anwesend. \\
    \bottomrule
    \end{tabular}}
    \caption{Translation from English to German. The words in color denote the copying tokens of which \textcolor{blue}{blue} denotes right copies and \textcolor{red}{\it red} denotes copying errors.}
    \label{tab:example}
\end{table}

\subsection{Copying Ratio}
\label{sec:cer}
\paragraph{Ratio} 
To measure the extent of the copying behaviors in NMT models, we calculate the ratio of copying tokens in translation outputs:
\begin{eqnarray}
\label{eq:ratio}
\mathrm{Ratio} = \frac{\sum_{i=1}^I\mathrm{count(copying\ token)}}{\sum_{i=1}^I\mathrm{count(token)}}
\end{eqnarray}
where $I$ denotes the total number of sentences in the test set. We count the number of ``copying token'' by comparing each input and output sentence pair. The denominator is the total number of tokens in output sentences. 
In general, higher Ratio values indicate more copying behaviors produced by the NMT model, and vice versa.

\begin{table}[t]
\centering
\scalebox{0.87}{
\begin{tabular}{lrrrrr}
\toprule
\multirow{2}{*}{} & \multicolumn{2}{c}{\bf Performance}  & \multicolumn{3}{c}{\bf Copying}   \\
\cmidrule(lr){2-3} \cmidrule(lr){4-6} 
 & \bf BLEU   &\bf TER   & \bf Ratio & \bf CER  &\bf \#S  \\
 \midrule
\sc Oracle & - & - &8.5\% &-   &9    \\ 
\random & 28.3   &60.7 &9.3\% &17.4  & 20    \\ 
\pre & 29.4  &59.4 &10.8\% &27.6  &50 \\
\bottomrule
\end{tabular}}
\caption{Results on the WMT14 En-De test set. {\sc Oracle} denotes the statistics on the reference. ``\#S'' denotes the number of instances whose overlaps between the source and target exceeding 50\%~\citep{ott2018analyzing}. Although \pre gains better model performance than \random, it also excessively copies tokens from the source.}
\label{tab:randomvsmbart}
\end{table}

\begin{figure*}[t] 
    \centering
    \includegraphics[width=0.9\textwidth]{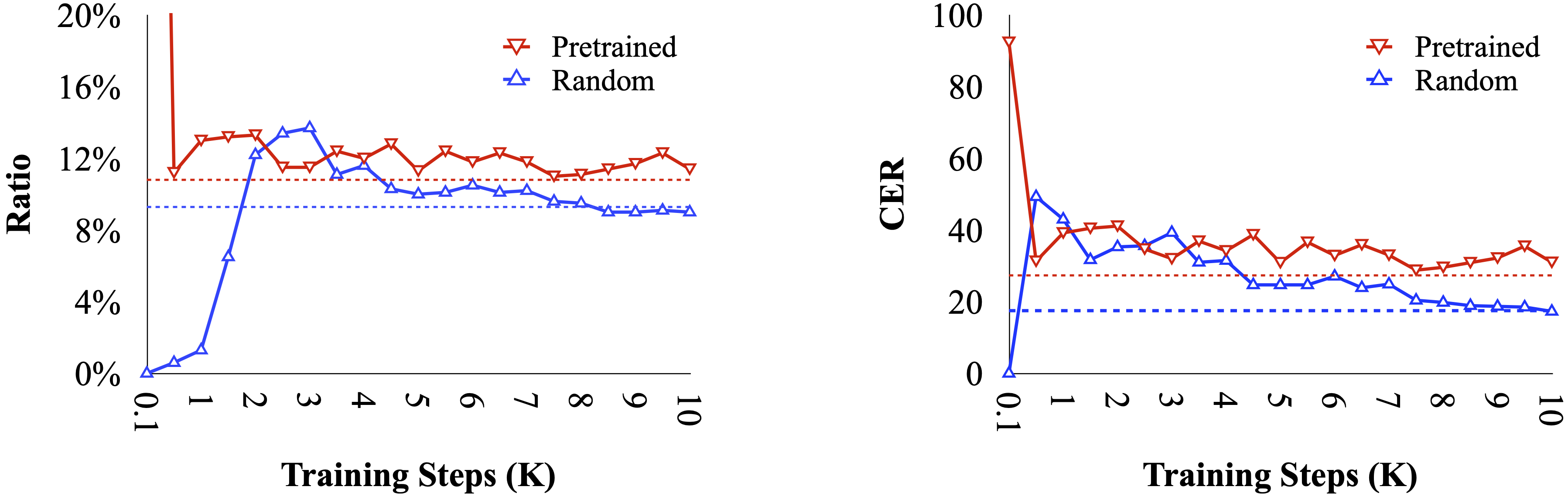}
    \caption{Copying ratio (left) and CER score (right) of \random and \pre at different training steps. Reference lines report the values of the final models gaining the lowest validation perplexities. \pre gains 89.4\% copying ratio at 0.1K step, which is omitted for better display clarity. \pre learns to forget its copying behaviors by reducing the copying ratio from 89.4\% to 10.8\% and CER from 92.4 to 27.6, while \random learns copying from scratch by increasing the copying ratio and CER from 0 to 9.3\% and 17.4, respectively.} 
    \label{fig:100-best}
\end{figure*}

\paragraph{Copying Error Rate (CER)} 
To further analyze the copying problem in NMT models, we propose to calculate the rate of incorrect copying tokens among all copied ones:
\begin{eqnarray}
\label{eq:cer}
\mathrm{CER} = \frac{\sum_{i=1}^I\mathrm{count(copying\ error)}}{\sum_{i=1}^I\mathrm{count(copying\ token)}}
\end{eqnarray}
where we count the number of ``copying error'' by checking whether the copying tokens are included in its reference sentence. The CER is expected to be zero, which indicates that all copying tokens are correct.
Table~\ref{tab:example} gives an example.
In experiments, Ratio and CER are computed based on words rather than sub-words. 
We further filter all punctuations, which are similar across different languages.

\paragraph{Model Performance}
We compare the performance and copying behaviors of the final models in Table~\ref{tab:randomvsmbart}.
The results show that although \pre improves the overall performance in terms of BLEU and TER scores, it tends to generate more copying errors, limiting its further improvement.
In the following part, we probe into the essence of the copying behaviors via carefully designed experiments.

\subsection{Learning Curves of Copying Behaviors}
\label{sec:curves}

We analyze copying behaviors in learning dynamics. Specifically, we translate the test set using intermediate checkpoints at different training steps, and then compute corresponding Ratio and CER values. We compare \random and \pre, and plot their learning curves in Figure~\ref{fig:100-best}.

\paragraph{Ratio} 
Two models behave quite differently in the early stages of training. Taking Step 100 for instance, \pre copies 89\% tokens while \random does not generate any copying tokens. 
This demonstrates that the copying habit in the pre-trained model is transferred to NMT models. 
As training proceeds, the copying behaviors of \pre are heavily suppressed, resulting in a rapid drop in Ratio. 
On the contrary, \random is able to quickly learn copying from scratch, leading to an upward trend. After learning curves become stable, \pre performs more copying behaviors than \random (10.8\% vs. 9.3\% Ratio).

\begin{table}[t]
\centering
\scalebox{0.81}{
\begin{tabular}{lrrrrr}
\toprule
\multirow{2}{*}{} & \multirow{2}{*}{\bf PPL}  & \multicolumn{2}{c}{\bf Rand$>$Pre} & \multicolumn{2}{c}{\bf Rand$<$Pre}   \\
    \cmidrule(lr){3-4}  \cmidrule(lr){5-6} 
 & \bf    & \bf Ratio& \bf CER  & \bf Ratio& \bf CER   \\
 \midrule
\random & 60.8   &10.0\% &18.5  & 8.7\%  &17.4    \\ 
\pre & 95.1  &9.5\% &14.9  &12.5\% &39.7      \\
\bottomrule
\end{tabular}}
\caption{Sentence perplexity on the test set. \pre's translation of worse perplexity (i.e., ``Rand$<$Pre'') also contains the higher Ratio (12.5\%) and CER (39.7) scores.}
\label{tab:ppl}
\end{table}

\paragraph{CER}
In general, the results of CER show similar trends to those observed in Ratio. 
In the beginning, the CER of \pre is extremely high (i.e., 92.4), revealing that most of the copying tokens are incorrect. 
The reason behind this phenomenon is that pre-trained models are accustomed to copying source words but the habit is overly transferred to the downstream translation models. 
The interesting finding is that \random also makes more mistakes on copying at the early training stage. 
Finally, the error rate of \pre is much higher than that of \random (27.6 vs. 17.4), showing that pre-trained models indeed expose harmful knowledge to NMT models.

Learning curves of two kinds of models perform in opposite ways: \random learns copying from scratch while \pre tries to forget this behavior. As a result, \pre copies more source tokens than \random and suffers severe copying errors. This motivates us to further investigate the effects of copying behaviors on NMT models in terms of translation quality.

\subsection{Effect of Copying Ratio}

\paragraph{Sentence Fluency}
\label{sec:cehurtfluency}
The copying tokens from the source usually contain some tokens that do not belong to the target language, which might hurt the fluency of generated translations.
Starting from this intuition, we use an external language model~\citep{ng2019facebook}\footnote{\url{https://dl.fbaipublicfiles.com/fairseq/models/lm/wmt19.de.tar.gz}} trained on in-domain data to evaluate the fluency of translation outputs.
As shown in Table~\ref{tab:ppl}, Random achieves a much better perplexity than mBART (60.8 vs. 95.1 PPL), demonstrating that the NMT model with pre-training generates less fluent sentences than that trained from scratch.

To take a closer look at the fluency gap, we divide outputs of each model into two subsets: sentences with better or worse perplexity by comparing \random and \pre. As seen, the fluency of translation is related to copying ratio and errors. The sentences with higher Ratio and CER scores tend to be less fluent. Taking \pre's worse subset for example (Rand$<$Pre), it gains a 12.5\% Ratio and 39.7 CER scores, confirming that excessive copying behaviors lead to negative effects in terms of translation fluency.

\begin{table*}[t]
\centering
\scalebox{0.81}{
\begin{tabular}{lrrrrrrrrrrrr}
\toprule
\multirow{2}{*}{} & \multicolumn{2}{c}{\bf Total}  & \multicolumn{2}{c}{\bf PROPN} & \multicolumn{2}{c}{\bf ADP} & \multicolumn{2}{c}{\bf NUM} & \multicolumn{2}{c}{\bf NOUN} &\multicolumn{2}{c}{\bf Others} \\ 
\cmidrule(lr){2-3} \cmidrule(lr){4-5} \cmidrule(lr){6-7} \cmidrule(lr){8-9} \cmidrule(lr){10-11} \cmidrule(lr){12-13}
&\bf Ratio &\bf CER$^{*}$  &\bf Ratio &\bf CER$^{*}$&\bf Ratio &\bf CER$^{*}$&\bf Ratio &\bf CER$^{*}$&\bf Ratio &\bf CER$^{*}$&\bf Ratio &\bf CER$^{*}$  \\
\midrule
\oracle & 8.5\% & -  &5.7\% &-   &1.1\%  &-  &0.8\% &- &0.5\%   &- &0.3\% &-   \\ \midrule
\random & 9.3\%& 17.4   &6.3\%&14.5   &1.3\%&25.8    &0.9\%&14.2   &0.5\%&21.3   &0.3\%&44.4  \\ 
\pre &10.8\% & 27.6    &7.5\%&27.3   &1.3\%&24.9    &0.9\%&14.0   &0.6\%&21.0   &0.5\%&63.8   \\ \hdashline 
\bf $\Delta$ &\bf +1.5\% &\bf  +10.2    &\bf +1.2\% &\bf +12.8   &0\% &-0.9   &0\% &-0.2  &+0.1\%  &+0.3  &+0.2\% &\bf +19.4 \\ 
\bottomrule
\end{tabular}}
\caption{Copying behaviors by part-of-speech (POS) bucket on the WMT14 En-De task. ``Oracle'' denotes the statistics in the reference. ``CER$^{*}$'' denotes only using the copying tokens belonging to each POS category for the CER calculation. $\Delta$ denotes the changes from the \random to \pre, in which the significant ones are {\bf bold}. Most of the copying tokens are found in translating proper nouns (PROPN) in \pre.}
\label{tab:ana-pos}
\end{table*}

\paragraph{Word Accuracy}
We also give a word-level analysis by bucketing copying tokens according to part-of-speech (POS) tags and calculate Ratio and CER in each type.
In experiment, we employ Stanford POS tagger with the \texttt{german-ud.tagger} model to automatically label output sentences~\cite{Toutanova:2003to}.
Table~\ref{tab:ana-pos} lists the results.
The ``Oracle'' (Row 1) denotes the statistics by comparing the source input and its reference.
As seen, most copying operations should occur in the type of proper noun (PROPN). This type occupies 5.7\% Ratios, followed by adposition (ADP), numeral (NUM), noun (NOUN), and other types (Others).

Compared with \random, we observe that the increase of Ratio for \pre mainly attributes to copying PROPN words (+1.2\%). In addition, \pre generates more copying errors (+10.2), especially on PROPN and Others types (+12.8 and +19.4). These results reveal that it is necessary to pay more attention to proper nouns on controlling copying behaviors for NMT.

\section{Controlling Copying Behaviors}
Based on the above experiments, we prove that pre-training indeed changes the copying behaviors of NMT models, hurting the sentence fluency and word accuracy of generated translations.
To alleviate this issue, we propose a simple and effective method {\it copying penalty} to make the copying behaviors in NMT controllable.

\subsection{Copying Penalty (CP)}
To control copying behaviors in NMT, an intuitive way is generating copying tokens only when the model is of high confidence.
To this aim, we propose to modify the probability distribution predicted by the NMT model, decreasing the predicting probability of the tokens also occurred in the source (i.e., weakening the model confidence of making copying predictions).
In this way, for those predictions are wavering between copying and translating, the model is more likely to translate them, and thus only those confident copying tokens will be retained.
Specifically, during inference, the predicting probability of $t$-th time step is as follows:
\begin{eqnarray}
\label{eq:vanillasearch}
P(y_{j} | y_{<j},\mathbf{x}) \in \mathds{R}^{\mathcal{V}} = \mathrm{softmax}(\mathbf{y}_t)
\end{eqnarray}
\noindent where $P(y_{j} | y_{<j},\mathbf{x})$ denotes the probability over the whole target vocabulary and $\mathbf{y}_t$ denotes the decoder output of $t$-th time step. 
The search algorithm (e.g., beam search) will take this probability distribution as a candidate to find the final translation of the source sentence.

Copying penalty regularizes the prediction probability of each time step by element-wisely multiplying a new constraint $\mathrm{CP} \in \mathds{R}^{\mathcal{V}}$:
\begin{eqnarray}
\label{eq:cp}
\mathrm{CP} = \left\{
            \begin{array}{lr}
             1, & y_{j}  \notin \mathbf{x}/ \mathcal{C}_\mathrm{punc}  \\
             \alpha , & y_{j}  \in \mathbf{x} / \mathcal{C}_\mathrm{punc}
             \end{array}
             \right.
\end{eqnarray}
\noindent where $\alpha$ is a hyperparameter to control the penalty which can be tuned on the development set, similar to length penalty~\citep{wu2016google}.
$\mathbf{x} / \mathcal{C}_\mathrm{punc}$ denotes the set of sources tokens excluding punctuation and \texttt{eos}, which means that the prediction probabilities of punctuation and \texttt{eos} will not be penalized.
For those predictions not belonging to the source, their probabilities keep the same.
But for those predictions that are copied from the source, their probabilities will be $\alpha$ times as large/small as before and the model will be more/less likely to choose them as candidates for searching.

The proposed method is simple and effective: 1) It does not change the model architecture and does not need any additional model training, thus no parameters needed to be newly introduced;
2) Its implementation only requires some low-cost matrix operations during model inference, slightly slowing the decoding speed;
and 3) It can significantly control the overall copying ratio of the model predictions, making the model accurately generate copying tokens, as shown in the following sections.

\begin{figure}[t]
\centering
\includegraphics[scale=0.29]{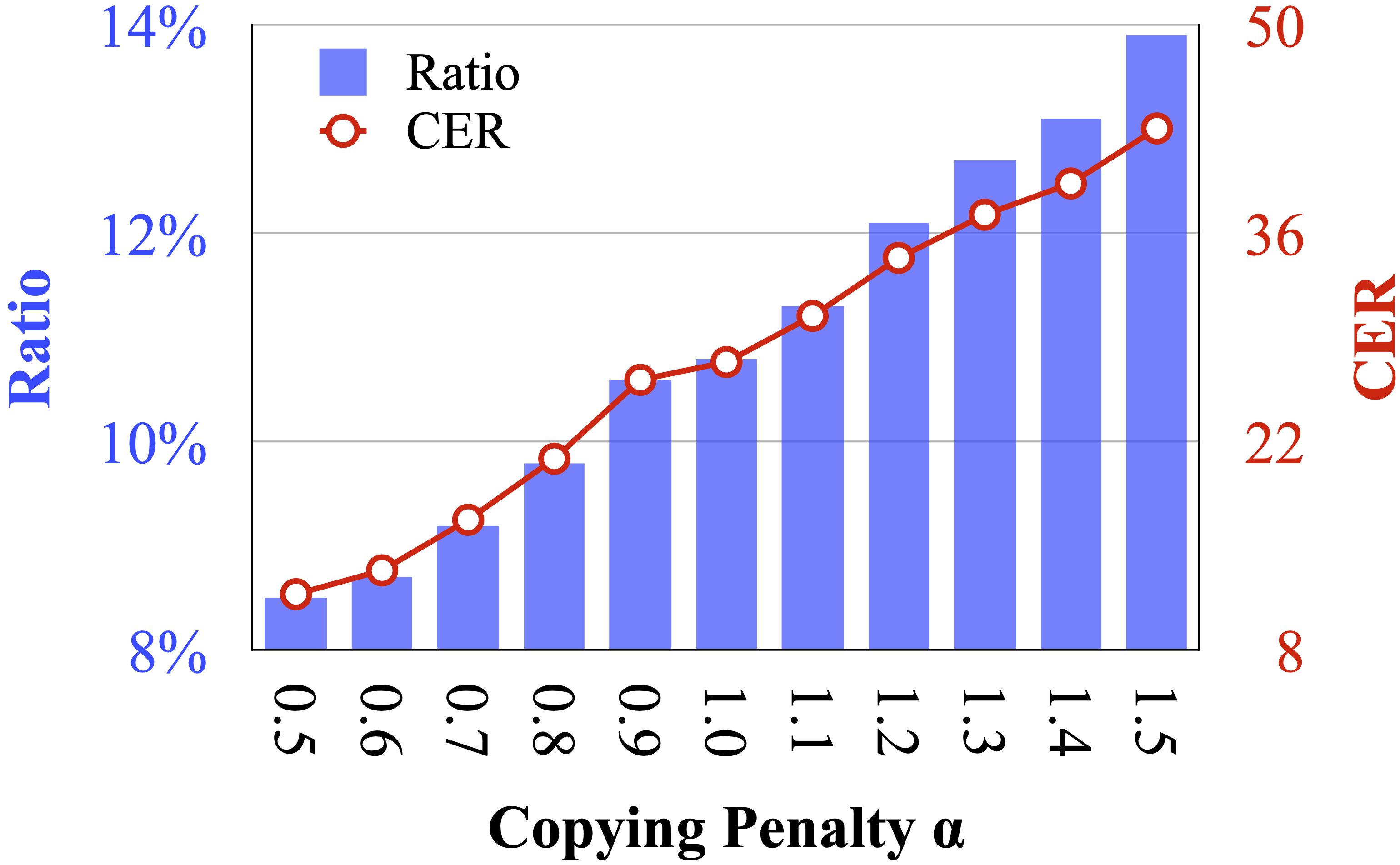}
\caption{Copying ratios and CER scores by different copying penalties in \pre. When setting CP smaller than 1 (i.e., penalizing copying), both the copying ratios and CER scores decrease, and vice versa.}
\label{fig:effect-of-cp}
\end{figure}

\paragraph{Effect of Copying Penalty}
Figure~\ref{fig:effect-of-cp} depicts the changes of the copying ratio and CER scores when setting CP to different values on the test data.
When setting CP smaller than 1 (i.e., punishing copying tokens), only those confident copying predictions will be made, and thus reducing both the copying ratio and CER scores.
Conversely, setting CP larger than 1 makes the model generate more copying tokens even some of them are of low confidence, thus both the copying ratios and CER scores increase.
Similar to length penalty~\citep{wu2016google}, we also tuned CP on the development data and found that setting CP to 0.7 wins the best BLEU score.
Therefore, we used this value for decoding test data in the following experiments.

Empirical results also show that the copying penalty is very efficient. 
When evaluated by a single 32GB V100 GPU card, the inference speed of \pre is about 612 token/s, and that of the model with CP is 607 token/s.
The extra latency of the copying penalty is negligible.

\begin{table}[t]
\centering
\scalebox{0.87}{
\begin{tabular}{lrrrrr}
\toprule
\multirow{2}{*}{} & \multicolumn{2}{c}{\bf Performance}  & \multicolumn{3}{c}{\bf Copying}   \\
\cmidrule(lr){2-3} \cmidrule(lr){4-6} 
 & \bf BLEU   &\bf TER   & \bf Ratio & \bf CER  &\bf \#S  \\
 \midrule
 \multicolumn{6}{l}{\bf All (3,003 sentences)} \\
\oracle & - & - &8.5\% &-   &9    \\ \midrule
\random & 28.3   &60.7 &9.3\% &17.4  & \bf 20    \\ 
\pre & 29.4  &59.4 &10.8\% &27.6  &50     \\ \hdashline 
\sc \quad+CP & \bf 29.6   &\bf 59.0  & \bf 9.2\% & \bf 16.8  & \bf 20     \\   
\midrule
\midrule
\multicolumn{6}{l}{\bf HasCopy (1,774 sentences)} \\
\oracle & - & -  &12.5\% &-  &9  \\ \midrule
\random &29.7  &59.4  &13.4\% &13.8  &\bf 15   \\ 
\pre &30.9  &57.8 &14.8\% &20.7  &32    \\ \hdashline 
\sc \quad+CP &\bf 31.0   &\bf 57.5 &\bf 13.0\%   &\bf 13.2 &17    \\   
\midrule
\midrule
\multicolumn{6}{l}{\bf NoCopy (1,229 sentences)} \\
\oracle & - & -  &0\% &-  &0  \\ \midrule
\random &25.4  &62.9  &\bf 1.2\%&-  &5   \\ 
\pre &26.2  &62.7 &2.7\%&-  &18    \\ \hdashline 
\sc \quad+CP &\bf 26.7  &\bf 62.0  &\bf 1.2\%&-  &\bf 3    \\   
\bottomrule
\end{tabular}}
\caption{Overall results on the WMT14 En-De test set. ``Oracle'' denotes the statistics on the reference. ``+CP'' denotes using the proposed copying penalty method for inference. ``\#S'' denotes the number of instances whose overlaps between the source and target exceeding 50\%. ``HasCopy'' denotes evaluating on the sampled test set containing copies between the source and target, while ``NoCopy'' denotes evaluating on the remained set without any copying. CER score is not applicable in NoCopy as all the copying tokens are copying errors.} 
\label{tab:main}
\end{table}

\subsection{Main Results}
Table~\ref{tab:main} lists the overall results of the model performance and copying behavior.
By looking at the part of evaluating the all test data of 3,003 sentences, the results first confirm the effectiveness of \pre that can consistently improve the model performance in terms of BLEU and TER scores.
However, the introduction of pre-trained knowledge also brings more copying properties to the model that increases the copying ratio, copying errors, and the number of copying sentences at the same time.
Thanks to the introduction of the copying penalty, the model successfully alleviates the copying errors (i.e., reducing the CER score from 27.6 to 16.8), making them be on par with \random, and thus further improve the BLEU and TER scores over the strong \pre.

\begin{table}[t]
\centering
\scalebox{0.87}{
\begin{tabular}{lcrrrr}
\toprule
\multirow{2}{*}{} &\multirow{2}{*}{\bf \#Num} & \multicolumn{2}{c}{\bf Total}  & \multicolumn{2}{c}{\bf PROPN}  \\ 
\cmidrule(lr){3-4} \cmidrule(lr){5-6} 
& &\bf Ratio &\bf CER   &\bf Ratio &\bf CER$^{*}$  \\
\midrule
All &\multirow{2}{*}{3,003} &10.8\%&27.6   &7.5\%&27.3   \\ 
\sc \quad+CP& &9.2\%&16.8  &6.3\%&15.1     \\ 
\hdashline 
Tgt-Ori&\multirow{2}{*}{1,500} &10.1\%&17.5    &6.6\%&15.6     \\ 
\sc \quad+CP & &9.4\%&12.6    &6.1\%&10.1     \\ 
\hdashline 
Src-Ori &\multirow{2}{*}{1,503} &11.4\%&34.4    &8.3\%&34.8     \\ 
\sc \quad+CP & &9.1\%&20.3    &6.4\%&18.9     \\ 
\bottomrule
\end{tabular}}
\caption{Copying behaviors of the source original and target original text in \pre. ``\#Num'' denotes the total number of sentences in each test set. The translation of source original text contains more copying tokens, and CP can reduce the copying ratio.}
\label{tab:copying-translationese}
\end{table}

To better understand how copying behaviors affect model performance, we split the test data into two subsets: HasCopy and NoCopy. 
One intuitive assumption is that copying errors would significantly hurt the performance of the NoCopy data since every copying token in the translation is a copying error.
The results confirm our assumption that \pre can only improve limited model performance on the NoCopy data (e.g., improving the TER scores from 62.9 to 62.7), but with the copying penalty, the copying errors less occur in \pre (i.e., reducing the copying ratio from 2.7 to 1.2 and copying sentences from 18 to 3) and thus, better model performance.

\paragraph{Sentence Fluency} 
The copying penalty improves sentence fluency.
In~\S\ref{sec:cehurtfluency}, we show that the perplexity of \pre (95.1) is worse than that of \random (60.9).
However, after introducing the copying penalty into \pre, the perplexity gets a significant drop from 95.1 to 62.3, which can be on par with \random.
This confirms our hypothesis that more copying behaviors hurt NMT in terms of translation fluency, and controlling copying behaviors can make the model generate fluent outputs.

\paragraph{Word Accuracy}
The copying penalty enhances the translation of PROPN.
As shown in Table~\ref{tab:copying-translationese}, the copying penalty improves the translations of proper nouns, reducing the copying ratio from 7.5\% to 6.3\% and the CER score from 27.3 to 15.1. 

To make headway into the translation of proper nouns, we further investigate the translations from various sources that usually show the large difference in the number of proper nouns~\citep{lembersky2012language}.
Specifically, we investigate the two kinds of sentence pairs in the WMT14 En-De test set: 1) the {\em source original text} (Src-Ori) that originated in English and was human-translated into German; 2) the {\em target original text} (Tgt-Ori) was translated in the opposite direction, originating in German with manual translation into English.

\citet{zhang2019effect} conclude that Tgt-Ori is artificially easier to translate, resulted in inflated scores for NMT models.
Our results of \pre positively support this conclusion that translating Tgt-Ori makes fewer copying errors and this might be a reason why it can win a better translation performance.
However, by looking at the last row, Src-Ori suffers from serious copying errors especially in translating proper nouns, making it harder to translate.
Encouragingly, the copying penalty nicely reduces the copying ratios and copying errors in translating both Src-Ori and Tgt-Ori.
These results further reveal the importance of controlling copying behaviors in NMT models since translating source original text is the core task of most NMT systems~\citep{graham-etal-2020-statistical}.

\begin{figure}[t]
\centering
\includegraphics[width=0.42\textwidth]{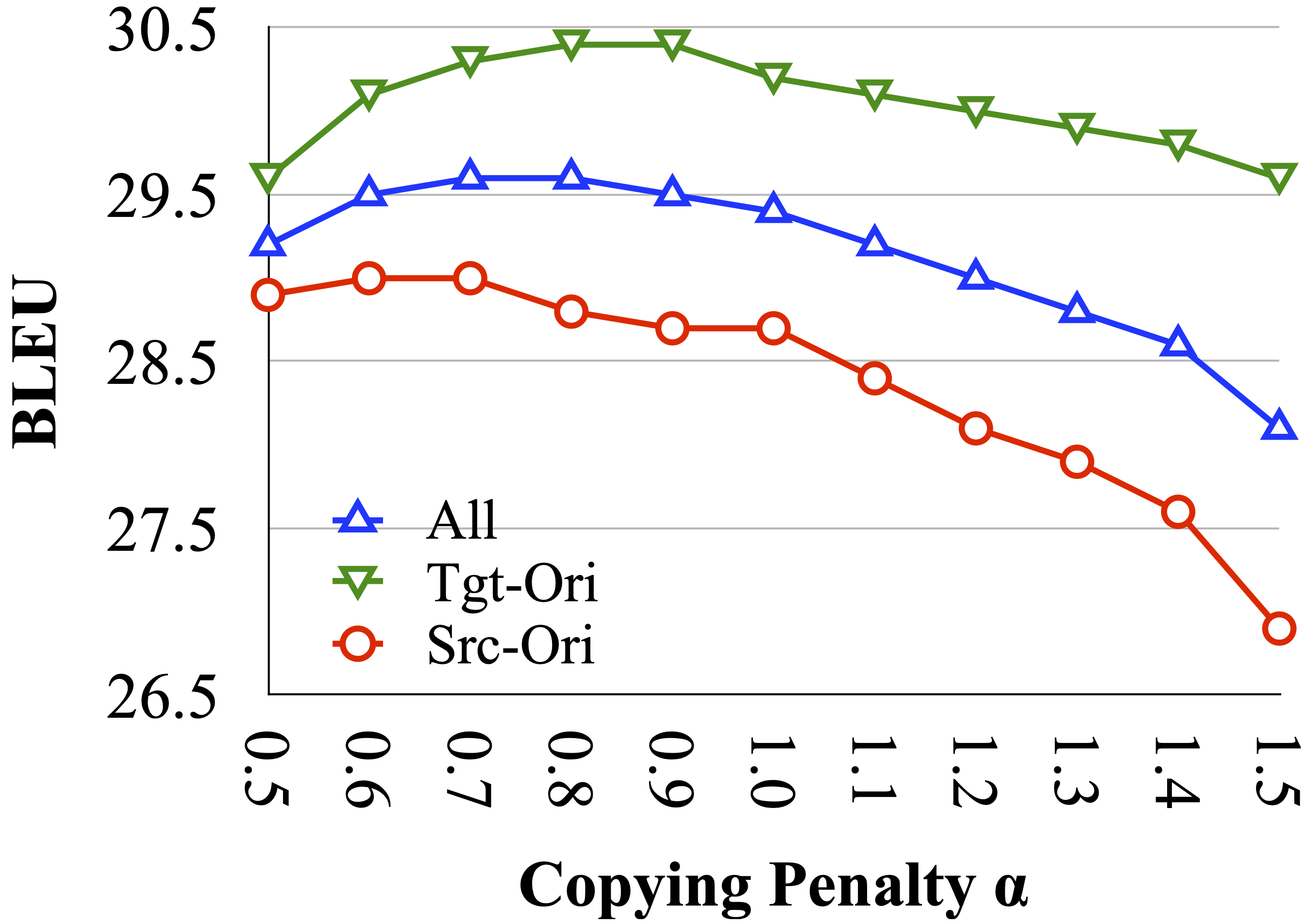}
\caption{BLEU scores of different copying penalties in \pre. Penalizing copying (i.e., $\alpha<1$ ) brings benefits to the translations of various sources. Translating source original sentences is more sensitive to copying behaviors, leading to a larger score degradation when encouraging copying (i.e., $\alpha>1$ ).}
\label{fig:effect-of-trans}
\end{figure}

The above results have shown that copying errors worsen the translation of Src-Ori.
To support the claim, we further investigate the effects of varying degrees of copying errors in the translations of Src-Ori and Tgt-Ori.
Figure~\ref{fig:effect-of-trans} shows the change of BLEU scores with different copying penalties.
Clearly, the translation of Src-Ori is more sensitive to copying errors and thus the BLEU scores get a sharp degradation when setting the copying penalty greater than 1, which verifies our claim.

\subsection{Out-of-domain Robustness}
Improving out-of-domain (OOD) robustness is one of the benefits of pre-training for NLP tasks~\citep{hendrycks2020pretrained,tu2020empirical}, but the OOD sentences usually contain some low-frequency proper nouns which are hard to translate~\citep{Ding2021ACL}.
In this part, we take the first step towards understanding how pre-training affects the OOD robustness of NMT models.

\paragraph{Setup} We followed~\citet{Muller:2019tq} to preprocess all the used data sets.\footnote{\url{https://github.com/ZurichNLP/domain-robustness/blob/master/scripts/preprocessing/preprocess_de_en.sh}}
We served the medical domain as the training domain (i.e., using the data from the medical domain for model training and validation), which consists of 1.1M training examples and 2,000 validation examples.
The test set of the medical domain contains 1,691 examples, while the test sets of the IT, Koran, law, and subtitle domains are with 2,000 examples respectively.
For training \random, we used the Transformer {\it base} setting with 32K batch size.
The model dropout is set to 0.3, while the dropouts for attention and inner-FFN activation are set to 0.2.
For \pre, apart from using 32K batch size, the other hyperparameters follow the training of the WMT14 English-German task.
The beam size is set to 5 and the length penalty is set to 1.4.
We evaluated the model performance on the OOD test sets from IT, Koran, Law, and Subtitles domains, and the averaged BLEU scores can be seen as the OOD robustness of each NMT model.

\begin{table}[t]
\centering
\scalebox{0.8}{
\begin{tabular}{lcrrrrr}
\toprule
\multirow{2}{*}{} & {\bf InD}  & \multicolumn{5}{c}{\bf OutD}  \\
\cmidrule(lr){2-2} \cmidrule(lr){3-7} 
&\bf Med. &\bf Avg.  &\bf IT &\bf Kor. &\bf Law &\bf Sub.  \\
\midrule
 \sc Existing  &  61.5 & 11.7 &17.1 &1.1 &25.3 &3.4 \\
\sc  \quad+Reg. & 60.8 & 13.1 &- &- &- &- \\
\hdashline 
\random   &  60.5 & 11.4 &20.5 &1.1 &20.5 &3.4    \\ 
\pre   &  63.1 & 17.6 &29.8 &2.4 &31.0 &7.3    \\ \hdashline 
\sc \quad+CP &  \bf 63.2 & \bf 18.3 &\bf 31.5 &\bf 2.5 &\bf 31.1 &\bf 7.9    \\ 
\bottomrule
\end{tabular}}
\caption{BLEU scores on the OPUS De-En translation task trained on the in-domain medical data. ``Existing'' and ``+Reg.'' denote the results of baseline and regularization method from~\citet{Muller:2019tq}. CP can significantly improve the translation performance of the IT domain that needs to copy more tokens from the source.}
\label{tab:copying-domain}
\end{table}

\paragraph{Results} Table~\ref{tab:copying-domain} lists the results.
Clearly, \pre substantially improves the performances of in-domain translation and OOD robustness, increasing the in-domain BLEU scores from 60.5 to 63.1 and the OOD BLEU scores from 11.4 to 17.6 respectively.
The copying penalty can further improve the OOD robustness of \pre that consistently improves the model performance of each OOD test set.
The copying penalty can even remarkably enhance \pre in translating the sentences from the IT domain (when setting CP to 1.2).
One possible reason is that the IT domain needs to copy more tokens from the source sentence than translating sentences from other domains, thus the copying penalty can play a greater role and bring a significant performance boost.
This also verifies the effectiveness of the copying penalty.

\section{Related Work}
\subsection{Pre-Training for NMT}
Recently, pre-training has been shown useful for transferring general knowledge to specific downstream tasks, including text classification, question answering and natural language inference~\citep{Peters:2018vk,radford2018gpt,Devlin:2018uk,liu2019roberta,yang2019xlnet}.
Compared with training from scratch, fine-tuning a pre-trained model on downstream datasets usually pushes state-of-the-art performances, while reducing computational and labeling costs.

Previous studies mainly investigate the effect of pre-training on NMT from two perspectives: 1) {\em knowledge extraction}, where a fixed pre-trained model is used to encode input sequences into features which are then fed into NMT models; and 2) {\em parameter initialization}, where part/all of the parameters of an NMT model are initialized by a pre-trained model and then training the model on downstream datasets (i.e., parallel corpus).

About knowledge extraction, \citet{yang2020towards} and \citet{Zhu2020Incorporating} explore enhancing encoder and decoder representations by leveraging pre-trained BERT models~\citep{Devlin:2018uk}. 
In addition, \citet{chen2020distilling} distill the soft labels from BERT to improve predictions for NMT.
These methods are effective but costly because the novel NMT architecture needed to be carefully designed and the computation graph has to store the parameters of both the pre-trained model and NMT model. 

About parameter initialization, pre-trained models in different architectures have been studied.
For the pre-trained model whose architecture is similar to Transformer encoder (e.g., BERT) or decoder (e.g., GPT~\citep{radford2018gpt}), the parameters of encoder and decoder can be independently initialized~\citep{lample2019cross,rothe2020leveraging}.
For the pre-trained model building upon the encoder-decoder architecture~\citep{sutskever2014sequence}, all the model parameters can be directly inherited by NMT, which is easy to use and effective~\citep{song2019mass,bart2020,lin2020pre,yang2020csp}.

In general, most previous works focus on designing novel pre-training methods and architectures to boost the model performance of NMT, but the understanding of pre-training for NMT is still limited.
This paper improves pre-training for NMT by first understanding its weakness in copying behavior, revealing the importance of further identifying the side-effect from pre-training.

\subsection{Copying Behaviors of NMT}
It is a common behavior in Seq2Seq models to copy source tokens to the target sentences, especially in monolingual generation tasks.
For example, \citet{gu2016incorporating} propose a copying mechanism to explicitly help model learn copying predictions, showing its effectiveness in the tasks of dialogue and summarization.

The copying behaviors also exist in NMT, particularly in languages that share some alphabets (e.g., English and German).
\citet{koehn2017six} observe that subword-based NMT~\citep{DBLP:journals/corr/SennrichHB16} outperforms statistical machine translation when translating/copying unknown words.
\citet{knowles2018context}~find that NMT is able to translate source words in specific contexts via copying (e.g., personal names followed by ``Mrs.''), and even these are unknown words.
However, too many copying signals (i.e., source and target sentences are identical) in training data may lead to one potential threat: NMT models prefer copying source tokens instead of translating them, resulting in performance degradation~\citep{ott2018analyzing,khayrallah2018copy}.

This paper broadens the understanding of copying behaviors in NMT models.
We observe that the translation of proper nouns in the source original text contains more copying tokens, which sheds light upon future works.

\section{Conclusion and Future Work}
We find that NMT models with pre-training are prone to generate more copying tokens.
We introduce a copying ratio and a copying error rate to quantitatively analyze copying behaviors in NMT evaluation. 
In addition, a simple and effective copying penalty is proposed to enhance the copying behaviors during model inference.
Experimental results prove the effectiveness of the copying penalty, which can effectively control copying behaviors and improve the overall model performance, especially for the domains (e.g., the IT) where much copying is needed.
Extensive analyses reveal that translating proper nouns in source original text generates more copying tokens, providing a direction for the following works on controlling copying behaviors of NMT models.

In the future, we would like to test the effectiveness of the copying penalty on the NMT models with other powerful pre-trained models, and explore more kinds of discrepancies between LM pre-training and NMT training which can be investigated to improve the performance of NMT
models. It is also worthwhile to adapt it to other Seq2Seq tasks that need to make a large number of copying predictions, e.g., text summarization and grammar error correction~\citep{liu2021understanding}.

\section*{Acknowledgement} 
This work was supported in part by the Science and Technology Development Fund, Macau SAR (Grant No. 0101/2019/A2), and the Multi-year Research Grant from the University of Macau (Grant No. MYRG2020-00054-FST). We thank the anonymous reviewers for their insightful comments.

\bibliography{acl2021}
\bibliographystyle{acl_natbib}

\end{document}